\documentclass[10pt,twocolumn,letterpaper]{article}

\usepackage{cvpr}              

\usepackage[dvipsnames]{xcolor}

\definecolor{cvprblue}{rgb}{0.21,0.49,0.74}
\usepackage[pagebackref,breaklinks,colorlinks,citecolor=cvprblue]{hyperref}
\usepackage[linesnumbered,ruled]{algorithm2e}
\usepackage{amsmath}
\usepackage{amsfonts}
\usepackage{multirow}
\usepackage{pifont}
\usepackage{newunicodechar}
\newunicodechar{✓}{\ding{51}}
\newunicodechar{✗}{\ding{55}}
\newcommand{\baby}{\textit{Iris}\xspace}
\def\paperID{10800} 
\def\confName{CVPR}
\def\confYear{2025}

\title{%
\baby: Breaking GUI Complexity with Adaptive Focus and Self-Refining%
}

\author{Zhiqi Ge$^{1}$, Juncheng Li$^{1}$, Xinglei Pang$^{1}$, Minghe Gao$^{1}$, Kaihang Pan$^{1}$\\ 
Wang Lin$^{1}$, Hao Fei$^{2}$, Wenqiao Zhang$^{1}$, Siliang Tang$^{1}$, Yueting Zhuang$^{1}$\\
Zhejiang University$^{1}$, National University of Singapore$^{2}$\\
}
\begin{document}
\maketitle
\begin{abstract}
    Digital agents are increasingly employed to automate tasks in interactive digital environments such as web pages, software applications, and operating systems. While text-based agents built on Large Language Models (LLMs) often require frequent updates due to platform-specific APIs, visual agents leveraging Multimodal Large Language Models (MLLMs) offer enhanced adaptability by interacting directly with Graphical User Interfaces (GUIs). However, these agents face significant challenges in visual perception, particularly when handling high-resolution, visually complex digital environments. This paper introduces Iris, a foundational visual agent that addresses these challenges through two key innovations: Information-Sensitive Cropping (ISC) and Self-Refining Dual Learning (SRDL). ISC dynamically identifies and prioritizes visually dense regions using a edge detection algorithm, enabling efficient processing by allocating more computational resources to areas with higher information density. SRDL enhances the agent's ability to handle complex tasks by leveraging a dual-learning loop, where improvements in referring (describing UI elements) reinforce grounding (locating elements) and vice versa, all without requiring additional annotated data. Empirical evaluations demonstrate that \baby achieves state-of-the-art performance across multiple benchmarks with only 850K GUI annotations, outperforming methods using 10x more training data. These improvements further translate to significant gains in both web and OS agent downstream tasks.
\end{abstract}
    
\section{Introduction}
\label{sec:intro}
\begin{figure}[t]
    \centering
    \begin{subfigure}{\columnwidth}
        \centering
        \includegraphics[width=\columnwidth]{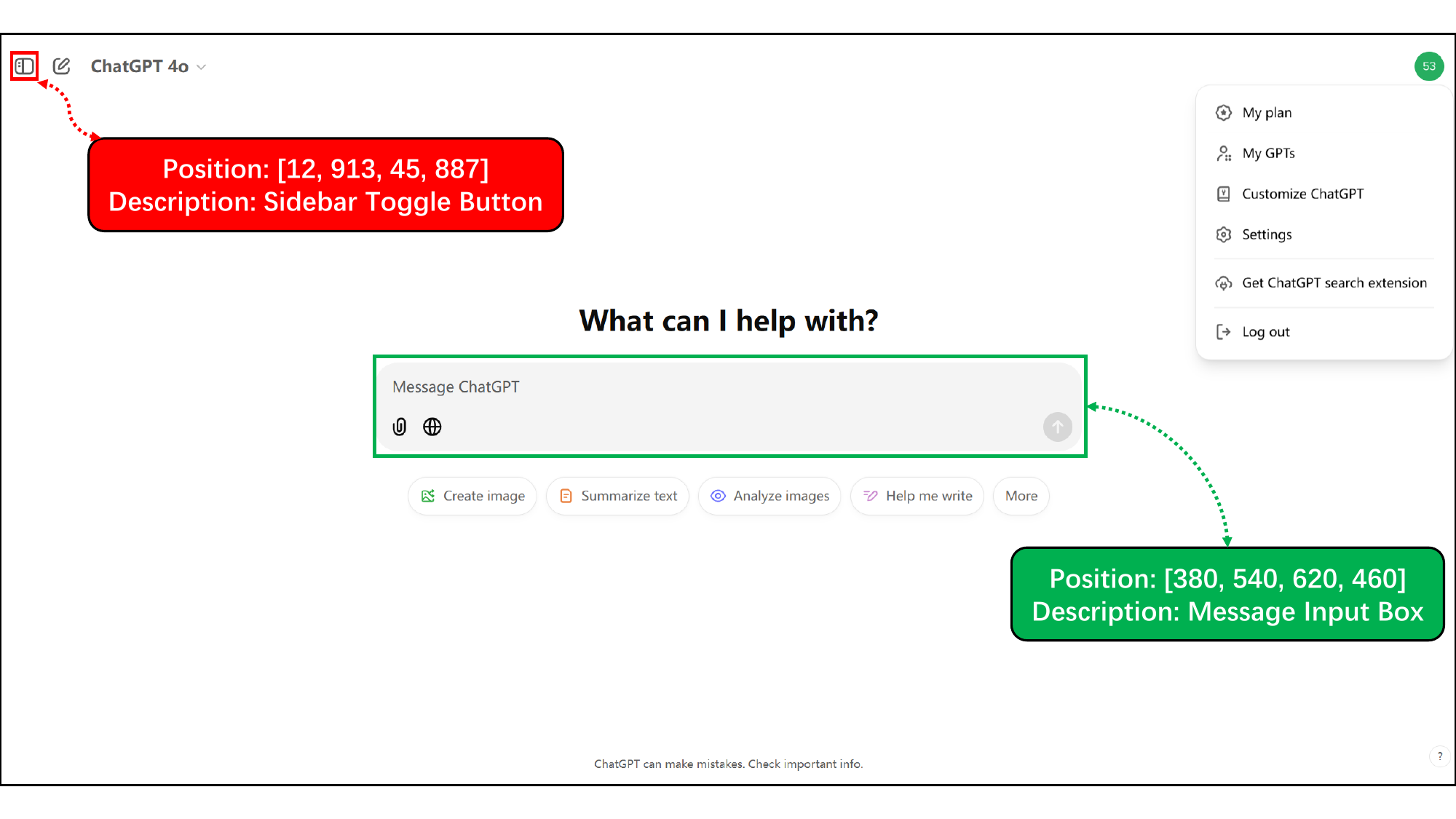}
        \caption{Example of GUI elements: Training data tend to focus on large, common elements such as input boxes (\textcolor{green}{green}) while real applications often involve small, uncommon elements such as sidebar buttons (\textcolor{red}{red}).}
        \vspace{0.5em}
        \label{fig:fig1a}
    \end{subfigure}
    \begin{subfigure}{\columnwidth}
        \centering
        \includegraphics[width=\columnwidth]{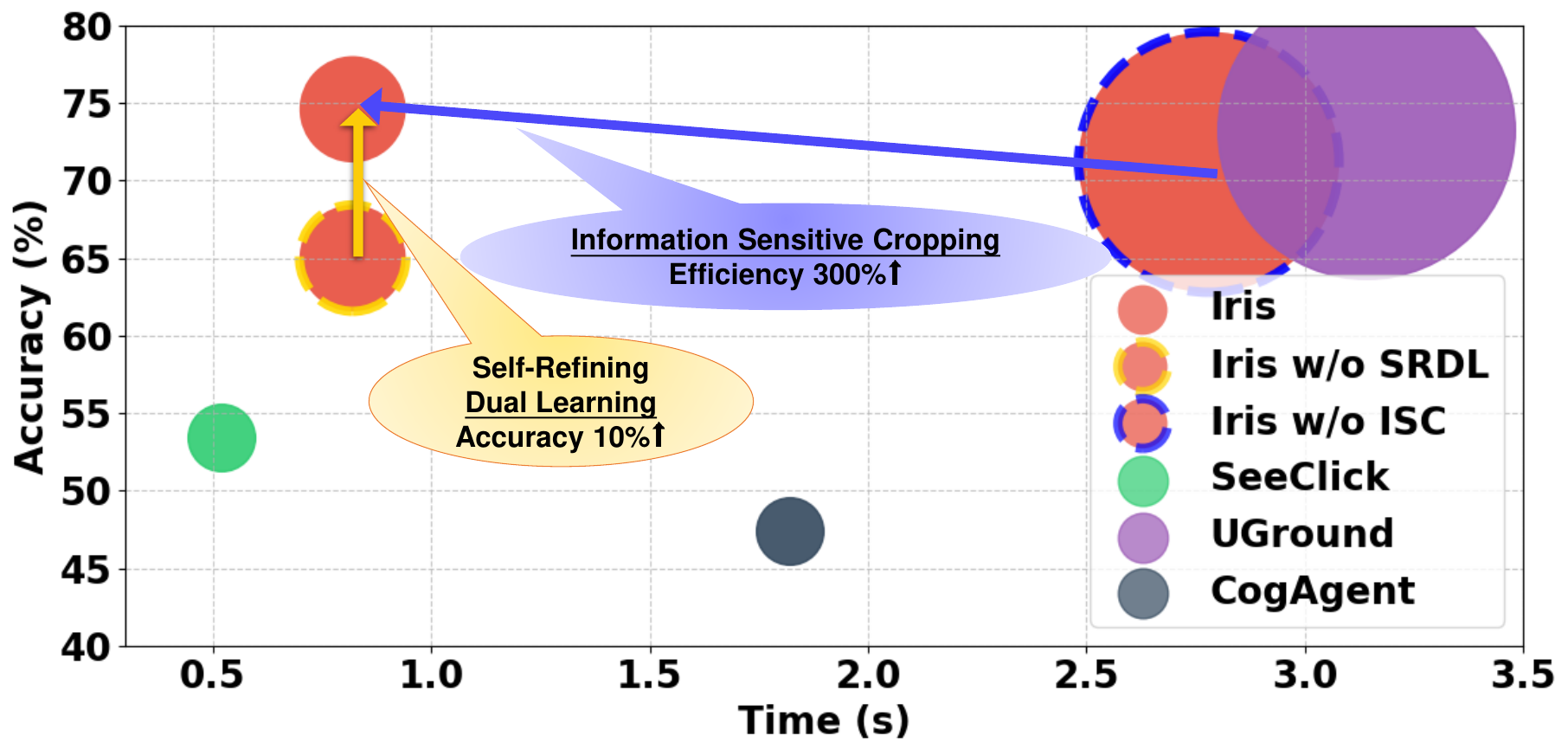}
        \caption{Performance analysis on ScreenSpot among different models: Bubble size indicates visual token count; arrows show ISC efficiency gain and SRDL accuracy improvement.}
        \vspace{-0.5em}
        \label{fig:fig1b}
    \end{subfigure}
 
    \caption{(a) Training data bias in GUI interfaces where annotations mainly cover visually prominent, functionally simple elements while ignoring small but interaction-critical components. (b) Our method achieves both better efficiency through ISC and higher accuracy through SRDL, with visual token count adapted to UI complexity.}
    \label{fig:fig1}
    \vspace{-1.5em}
 \end{figure}

Digital agents \cite{hong2024cogagent,wang2024mobile,zhang2024ufo,you2024ferretui,gao2024generalist} are designed to automate tasks within interactive digital environments, such as web pages \cite{deng2024mind2web,wu2023webui,liu2024visualwebbench}, software applications \cite{lu2024gui,deka2017rico}, and operating systems \cite{rawles2024androidinthewild,sun2022meta,wang2024mobile}. These agents leverage environmental information and user intent to offer both indirect assistance (e.g., answering questions) and direct support (e.g., executing actions). Traditional textual agents \cite{wu2024copilot,deng2024mind2web,zhou2023webarena}, built on Large Language Models (LLMs) \cite{achiam2023gpt,touvron2023llama,touvron2023llama2,team2023gemini}, depend on predefined APIs or code snippets such as HTTP requests or structured data formats (HTML/XML). Since APIs and code vary across platforms and evolve over time, maintaining these agents requires frequent updates, limiting their scalability. In contrast, visual agents \cite{hong2024cogagent,you2024ferretui,cheng2024seeclick,gao2024fine}, powered by Multimodal Large Language Models (MLLMs) \cite{achiam2023gpt,liu2024llava,bai2023qwenvl,team2023gemini,li2023finetuning,fei2024enhancing,fei2024vitron}, show promise in becoming more versatile. By mimicking human interaction methods—perceiving the Graphical User Interface (GUI) and performing actions within it—these agents can naturally achieve compatibility across a wide range of platforms.

While the GUI as a universal interface simplifies the development of versatile digital agents, it simultaneously introduces significant challenges in both visual perception and functional understanding. These challenges appear in common GUI interfaces where visually prominent elements with simple functions coexist with visually subtle ones that require complex interactions. For instance, \cref{fig:fig1a} shows how current training data focuses on large input boxes with straightforward text entry functions, while missing critical UI components like small menu buttons that control complex navigation hierarchies. When handling such cases, agents often fail at both visual detection due to the elements' subtle appearance and task execution due to their complex functionalities. These difficulties can be attributed to two key factors:
\begin{itemize}
    \item \textbf{Architectural Limitations in Processing Heterogeneous GUI Information:} Digital tasks often involve high-resolution interfaces (e.g., 1920x1080) that exhibit significant heterogeneity - combining densely packed UI elements with large, empty background areas. This uneven distribution of information creates dual challenges: visually, models struggle to maintain both fine-grained perception of dense regions and contextual awareness of the entire screen; functionally, they face difficulties in understanding complex dependencies between elements, especially within a limited computational budget. Current model architectures lack the flexibility to efficiently allocate computational resources based on this heterogeneous nature of GUI interfaces.
    \item \textbf{Annotation-Biased Training for GUI Understanding:} The training process of digital agents often relies on annotated data that favors simple patterns - both visually (large, prominent UI components) and functionally (common operations like 'OK' or 'Cancel' buttons). This leads to limited understanding of both complex visual layouts and intricate functional relationships. Moreover, obtaining comprehensive annotations that cover both dense information regions and their functional interactions requires substantial manual effort and domain expertise, constraining the scalability of visual agents.
\end{itemize}

To address these challenges, in this paper, we introduce a foundational visual agent focused on GUI understanding and grounding, with \textbf{I}teratively self-\textbf{R}efining and \textbf{I}nformation-\textbf{S}ensitive understanding capabilities, named \baby. Our approach addresses the challenges from two perspectives: 1) From the architectural perspective, \baby introduces an information-sensitive model design that efficiently processes high-resolution UI images with varying complexities, enhancing both visual perception and functional understanding; 2) From the training perspective, \baby develops a dual-learning strategy that iteratively refines the model's visual and functional knowledge using unlabeled data, enabling comprehensive GUI understanding without extensive annotations. To implement the information-sensitive architecture, we first propose \textbf{Information-Sensitive Cropping} (ISC), a dynamic cropping method that adapts to the distribution of visual information on the screen. Unlike previous methods \cite{xu2024llavauhd,dong2024internlm4k,you2024ferretui} that segment images into equally sized sub-images and allocate the same number of tokens to each part, ISC employs a fast edge detection \cite{canny1986computational} algorithm to identify regions with high information density and selectively crop them. These cropped regions are enlarged, allowing the agent to allocate more tokens to represent detailed areas. Conversely, low-information areas are encoded with fewer tokens, reducing computational overhead. By balancing the token allocation based on the information content of each region, ISC enables the agent to efficiently process both high-resolution context and fine-grained details, enhancing its visual perception capabilities.

To enhance \baby's robustness in challenging scenarios, we introduce a \textbf{Self-Refining Dual Learning} (SRDL) strategy that leverages the synergistic relationship between two complementary tasks: \textit{referring} (generating UI element descriptions) and \textit{grounding} (locating elements from descriptions). This dual approach creates a positive feedback loop—improvements in referring lead to better understanding of visual and functional characteristics, which enhances grounding accuracy, while more precise grounding provides richer spatial and contextual information that yields more accurate descriptions. A key advantage of SRDL is its ability to autonomously identify and learn from challenging elements without additional annotated data, addressing the common training bias toward easily identifiable UI components. The process begins by detecting difficult cases where \baby struggles, identified through either visual complexities (revealed by ISC information patterns) or functional challenges (indicated by historical performance). For these cases, \baby initiates a dual-learning cycle: it generates element descriptions and attempts to locate them based on these descriptions, using any inconsistencies between predicted and actual locations as feedback for iterative refinement. This self-reinforcing process not only enhances \baby's ability to handle complex UI tasks independently but also develops a deeper understanding of UI characteristics that transcends simple object detection while reducing dependence on labeled training data.

Empowered by ISC and SRDL, \baby demonstrates that enhanced foundational GUI understanding capabilities directly improve performance on complex sequential decision-making tasks in multimodal agent scenarios. As shown in \cref{fig:fig1b}, our method achieves significant improvements through complementary contributions: ISC delivers a 300\% efficiency gain by reducing processing time from 3s to 1s, while SRDL enhances accuracy by 10\% through better handling of challenging UI elements. Using the same training data as SeeClick \cite{cheng2024seeclick}, \baby achieves substantially better performance, improving accuracy from 53\% to 75\%. This performance matches UGround \cite{gou2024navigating}, which uses more than 10x the amount of training data, while requiring only one-third of its processing time. These efficiency and accuracy gains translate to consistent improvements across downstream tasks, with significant gains on both web agent \cite{deng2024mind2web} and OS agent \cite{rawles2024androidinthewild} benchmarks, indicating that \baby's enhanced GUI understanding capabilities are applicable to a wide range of digital agent scenarios.
\section{Related Work}
\label{sec:related}
\subsection{Visual Digital Agents} Recent works of visual digital agents have demonstrated significant progress in multi-platform capabilities, with systems like CogAgent \cite{hong2024cogagent}, SeeClick \cite{cheng2024seeclick} and UGround \cite{gou2024navigating} enabling navigation across both PC web pages and Android devices. Notably, SeeClick introduced an innovative vision-only approach that relies solely on screenshots for GUI interaction, enhanced through specialized GUI grounding pre-training. This eliminated the need for structured data like HTML that previous agents required, while achieving strong performance across mobile, desktop and web interfaces. Their work also established GUI grounding as a fundamental capability for visual agents, demonstrating its direct correlation with improved performance on downstream automation tasks. The development of comprehensive benchmarks like ScreenSpot \cite{cheng2024seeclick} and GroundUI \cite{zheng2024agentstudio} has facilitated evaluation of these agents' capabilities across diverse digital environments. 

\subsection{High Resolution MLLMs}
Recent work has made significant progress in handling high-resolution images in MLLMs, particularly crucial for GUI interfaces that often exceed 1000px. Approaches like AnyRes \cite{xu2024llavauhd,li2024ferret2,liu2024sphinx,dong2024internlm} tackled this by splitting images into 336x336 or 448x448 grids, though this could impact efficiency and contextual understanding across grid boundaries. More recent architectures like Qwen2-VL \cite{wang2024qwen2} introduce dynamic resolution support through innovations such as 2D-RoPE and multimodal position embedding, allowing flexible processing of varying image sizes while maintaining positional information. Despite these methods' ability to handle high-resolution inputs, they share a common limitation of treating all pixels and regions equally, leading to significant efficiency issues when processing GUI images where information is unevenly distributed. In contrast, our approach dynamically adapts computational resources based on information density, enabling more efficient processing of complex GUI interfaces.

\section{Method}
\label{sec:method}

\begin{figure*}[t]
    \centering
    \includegraphics[width=\textwidth]{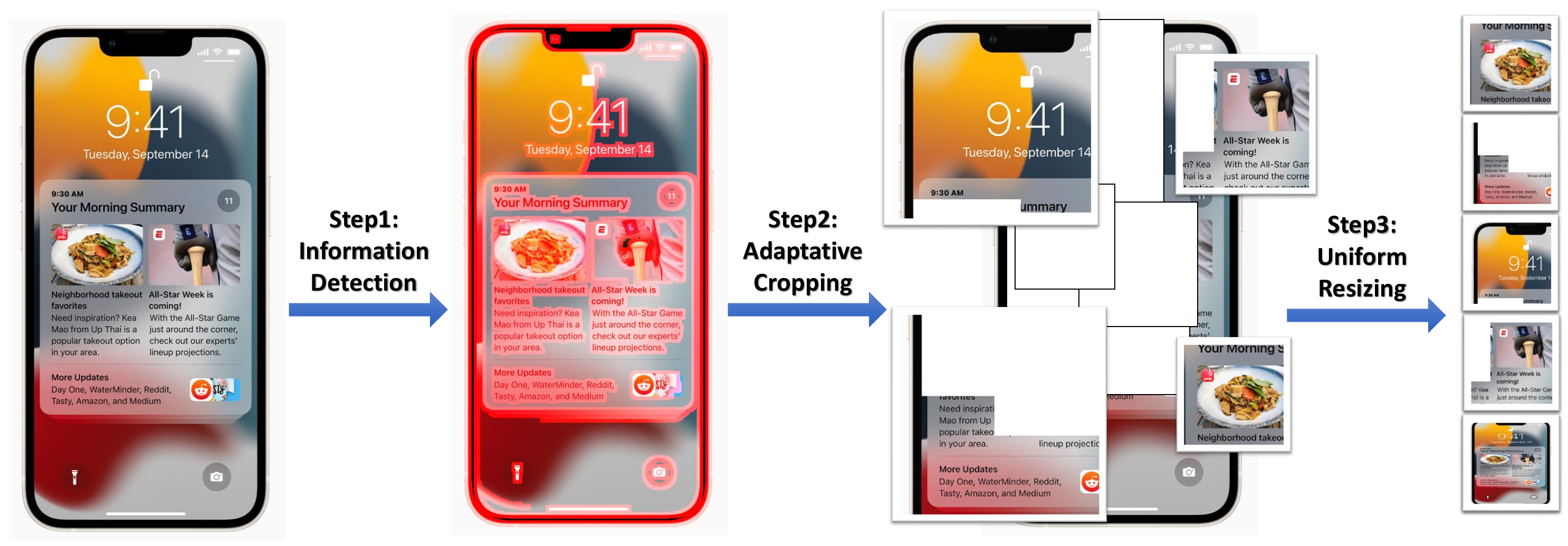}
    \caption{Illustration of the Information-Sensitive Cropping process. The original high-resolution screen image is segmented into smaller sub-images based on the distribution of visual information, ensuring that each sub-image captures a balanced amount of meaningful content.}
    \label{fig:isc}
\end{figure*}

In this section, we present the design and implementation details of \baby, a foundational visual agent focused on GUI understanding and grounding. We first introduce the preliminary concepts and task formulation in \cref{sec:preliminary}, setting the foundation for the referring and grounding tasks. We then describe two core components that distinguish \baby: \textbf{Information-Sensitive Cropping (ISC)} in \cref{sec:isc}, which dynamically segments high-resolution screens for efficient processing, and \textbf{Self-Refining Dual Learning (SRDL)} in \cref{sec:srdl}, which iteratively enhances the model's performance without external annotations.

\subsection{Preliminary and Task Formulation}
\label{sec:preliminary}

Our approach focuses on enhancing GUI understanding through two complementary tasks: referring and grounding. Each UI element is characterized by two key attributes:

\noindent\textbf{Position and Description.} A UI element's position is defined by its bounding box coordinates \(\mathbf{p} = (x_1, y_1, x_2, y_2)\), where \((x_1, y_1)\) and \((x_2, y_2)\) denote the top-left and bottom-right corners. The element's description \(D\) encompasses both its visual appearance (e.g., element type, displayed text) and functional role in the interface.

Based on these attributes, we formulate two core tasks:

\noindent\textbf{Referring:} Given a screen image and position \(\mathbf{p}\), generate a comprehensive description \(D\) of the UI element at that location. Let \(R(\cdot)\) denote the referring function:
\begin{equation}
    D = R(\mathbf{p})
\label{eq:referring}
\end{equation}

\noindent\textbf{Grounding:} Given a screen image and description \(D\), locate the corresponding UI element's position. Let \(G(\cdot)\) denote the grounding function:
\begin{equation}
    \mathbf{p} = G(D)
\label{eq:grounding}
\end{equation}

Together, referring and grounding serve as the core tasks in the development of digital agents that can effectively understand and interact within GUI environments.

\subsection{Information-Sensitive Cropping}
\label{sec:isc}

The core goal of Information-Sensitive Cropping (ISC) is to dynamically segment a high-resolution screen image into smaller, variable-sized sub-images based on the distribution of visual information. This adaptive approach ensures that each sub-image captures a relatively balanced amount of meaningful information, avoiding the pitfalls of uniform cropping strategies that can either neglect critical details or waste computational resources on irrelevant regions. The ISC process is structured into three key steps: information detection, adaptive cropping, and uniform resizing. Each step is described below.

\noindent\textbf{Information Detection.} We employ edge detection to identify visually significant regions, leveraging the observation that meaningful GUI elements typically have distinctive boundaries. Given an input image \(I\), we generate an information indication matrix \(M \in \{0, 1\}^{n \times m}\), where \(M_{i,j} = 1\) indicates the presence of meaningful visual information. See Appendix \cref{sup:isc} for implementation details.

\noindent\textbf{Adaptive Cropping.}
Given the edge detection matrix $M$, we employ a multi-scale sliding window approach to identify and extract information-rich regions. Let:

\begin{itemize}
    \item $k_t$: Current window size at iteration $t$
    \item $\rho_{\text{min}}$: Base density threshold
    \item $\alpha$: Window expansion factor
    \item $\Omega$: Set of extracted sub-images, where each element is $(x, y, k, \text{id}, \text{density})$
\end{itemize}

The process starts with a minimum window size $k_{\text{min}}$ and iteratively expands it by factor $\alpha$. At each scale, we:\\
1) Slide the window with step size $\text{step} = \max(k_t/4, 32)$\\
2) Calculate edge density for current window\\
3) Extract regions where density exceeds $\rho_k = \rho_{\text{min}}/(k_t/k_{\text{min}})^2$\\
4) Mark processed regions to avoid overlap.
The detailed algorithm is shown in \cref{alg:adaptive-cropping}.
\begin{algorithm}[ht]
    \SetAlgoLined
    \KwIn{
        $M$: Edge matrix,
        $k_{\text{min}}$: Initial window size,
        $\rho_{\text{min}}$: Base threshold,
        $\alpha$: Expansion factor,
        $N_{\text{max}}$: Max regions
    }
    \KwOut{$\Omega$: Selected regions}

    $k \gets k_{\text{min}}$, $\Omega \gets \emptyset$\;
    
    \While{$k \leq \max(\text{height}, \text{width})$ \textbf{and} $|\Omega| < N_{\text{max}}$}{
        $\text{step} \gets \max(k/4, 32)$\;
        $\rho_k \gets \rho_{\text{min}}/(k/k_{\text{min}})^2$\;
        
        \For{$(x,y)$ with step $\text{step}$}{
            \If{$x + k \leq \text{width}$ \textbf{and} $y + k \leq \text{height}$}{
                $\text{density} \gets \sum M[y:y+k, x:x+k] / k^2$\;
                
                \If{$\text{density} \geq \rho_k$}{
                    $\text{id} \gets |\Omega| + 1$\;
                    Add $(x, y, k, \text{id}, \text{density})$ to $\Omega$\;
                    $M[y:y+k, x:x+k] \gets 0$\;
                }
            }
        }
        $k \gets \lceil \alpha k \rceil$\;
    }
    
    Sort $\Omega$ by density\;
    \Return{$\Omega$}

    \caption{Adaptive Region Extraction}
    \label{alg:adaptive-cropping}
\end{algorithm}

\noindent\textbf{Uniform Resizing.}
Following the cropping step, we get a set of sub-images \( \Omega \), where each sub-image varies in size but contain a balanced amount of visual information. To ensure every visual token convey meaningful information, we down scale each sub-image to a fixed size (e.g., \(224 \times 224\), depending on visual encoder) using bilinear interpolation, then sending them to MLLM for further processing.

\noindent\textbf{Computational Efficiency.} The ISC process is remarkably efficient, typically requiring less than 0.1 seconds on CPU, and can run in parallel with GPU operations without introducing additional latency to the pipeline. See Appendix \cref{sup:isc_efficiency} for detailed performance analysis.

\noindent\textbf{Summary.} Through the ISC mechanism, \baby can efficiently process high-resolution screens by focusing on the most informative regions, meanwhile reducing attention on irrelevant areas, thereby enhancing its performance in both speed and accuracy.

\subsection{Self-Refining Dual Learning}\label{sec:srdl}

\begin{figure}[t]
    \centering
    \includegraphics[width=\columnwidth]{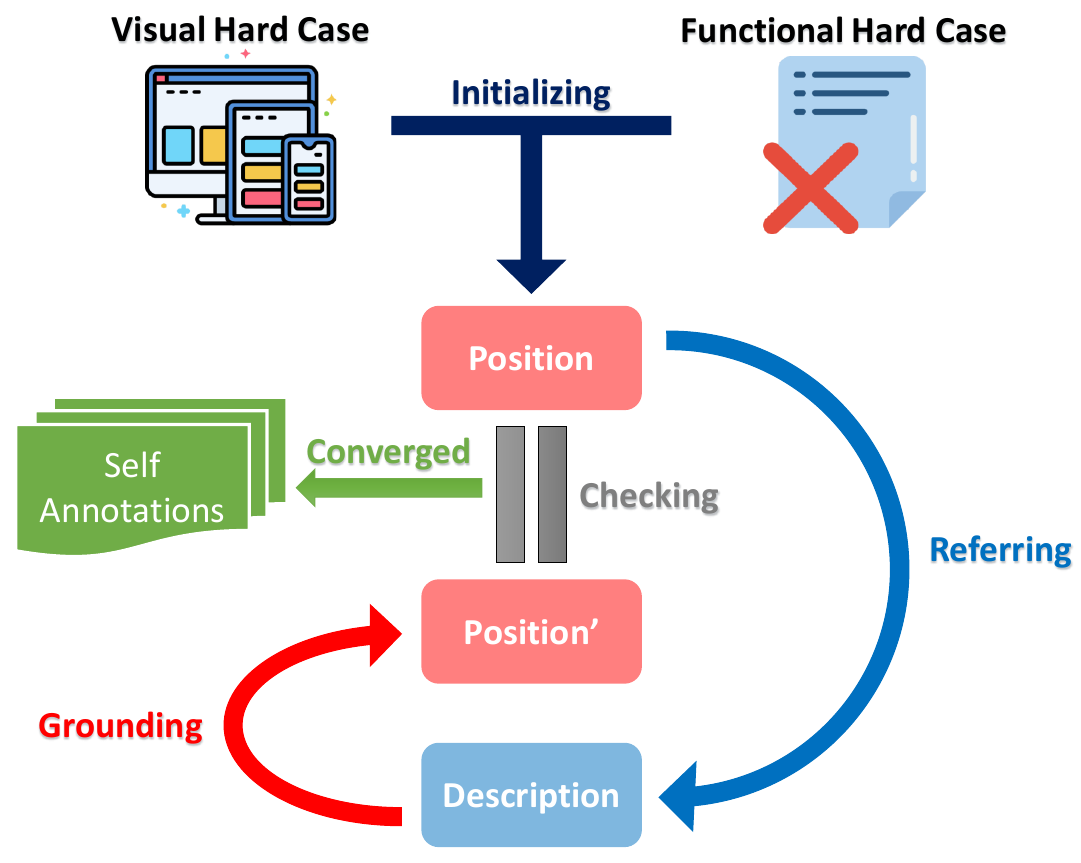}
    \caption{Illustration of the Self-Refining Dual Learning (SRDL) workflow. Starting from visual and functional hard cases, the system iteratively refines performance through a dual-learning loop of referring and grounding tasks, generating self-annotations when position predictions converge.}
    \label{fig:srdl}
\end{figure}

This section introduces \textbf{Self-Refining Dual Learning}, which allows \baby\ to autonomously discover and learn from difficult samples without relying on additional annotations. By leveraging the dual nature of referring and grounding tasks, the agent engages in self-play, iteratively refining its performance through a synergy-driven learning loop.

\noindent\textbf{Dual-Learning Loop for Referring and Grounding.} The core idea of the dual-learning loop is to exploit the complementary nature of referring (description generation) and grounding (element localization). Given a GUI image, \baby\ first enumerates all UI elements by prompting itself with \textit{"What are the UI elements in this image?}", treating each element's name as its \textit{basic description}. For each element, it performs \textbf{grounding} to locate the element's \textit{position} \( \mathbf{p} \). It then performs \textbf{referring} to re-generate the description from the grounded position. 

Given that a UI element's description can vary, we focus on the consistency of its position \( \mathbf{p} \) to determine convergence. If the grounded position from successive iterations remains stable, the output is considered converged, and the resulting sample can be added to the training set.

The dual-learning loop can be represented as follows. Let \( R(\cdot) \) denote the referring function, \( G(\cdot) \) the grounding function, \( D \) a description, and \( \mathbf{p} \) a position. The objective is to ensure that:

\begin{equation}
\operatorname{Sim}\left(G\left(R\left(\mathbf{p}\right)\right), \mathbf{p}\right) > \tau
\label{eq:dual-learning}
\end{equation}

where \( \operatorname{Sim}(\cdot, \cdot) \) is a similarity function (e.g., Intersection over Union, IoU) used to compare positions, and \( \tau \) is the convergence threshold. If the similarity exceeds the threshold, the sample is treated as a self-annotated example and added to the training phase.

\begin{algorithm}
    \SetAlgoLined
    \KwIn{GUI image \( I \), similarity threshold \( \tau \), max iterations \( N_{\text{max}} \)}
    \KwOut{Self-annotated samples \( \mathcal{S} \)}
    Initialize \( \mathcal{S} \gets \emptyset \)\;
    \For{each UI element \( i \) in \( I \)}{
        \( D_i \gets \text{Basic Description of UI element } i \)\;
        \( \mathbf{p}_i \gets G(D_i) \) 
        \( D'_i \gets R(\mathbf{p}_i) \) 
        Initialize iteration counter \( n \gets 1 \)\;
        \While{\( \operatorname{Sim}(\mathbf{p}_i, G(D'_i)) \leq \tau \) \textbf{and} \( n \leq N_{\text{max}} \)}{
            \( \mathbf{p}_i \gets G(D'_i) \) 
            \( D'_i \gets R(\mathbf{p}_i) \) 
            \( n \gets n + 1 \)\;
        }
        \If{\( \operatorname{Sim}(\mathbf{p}_i, G(D'_i)) > \tau \)}{
            \( \mathcal{S} \gets \mathcal{S} \cup \{(D_i, \mathbf{p}_i)\} \)\;
        }
    }
    \Return \( \mathcal{S} \)
    \caption{Dual-Learning Loop for Referring and Grounding}
    \label{alg:dual-learning-iterative}
\end{algorithm}

While effective, this strategy may concentrate learning on simpler samples and overlook difficult cases. Furthermore, it does not account for the model's historical performance, limiting its ability to address specific weaknesses. To overcome these limitations, we introduce two targeted hard case mining methods: Visual Hard Case Mining and Functional Hard Case Mining.

\noindent\textbf{Visual Hard Case Mining.} Visual hard cases are identified by analyzing the \textit{information matrix} \( M \) obtained from the ISC process. We use \textbf{spectral entropy}, derived from Fourier transform analysis, to quantify the density and complexity of visual information. Higher spectral entropy indicates dense visual content, which often corresponds to more challenging tasks for the agent.

Given an information matrix \( M \in \{0, 1\}^{n \times m} \), we first compute its 2D discrete Fourier transform (DFT):

\begin{equation}
F(u, v) = \sum_{i=0}^{n-1} \sum_{j=0}^{m-1} M(i, j) \cdot e^{-2\pi i \left( \frac{ui}{n} + \frac{vj}{m} \right)}
\label{eq:fourier}
\end{equation}

where \( F(u, v) \) is the Fourier spectrum representing the frequency components of the matrix. We then shift the zero-frequency component to the center using the `fftshift` operation for better interpretability. 

The spectral energy of each frequency component is given by \( |F(u, v)|^2 \). To quantify the distribution of energy across different frequencies, we define the \textbf{spectral entropy} \( H \) as:

\begin{equation}
H = - \sum_{k} p_k \log(p_k), \quad p_k = \frac{|F(u, v)|^2}{\sum_{u,v} |F(u, v)|^2}
\label{eq:spectral-entropy}
\end{equation}

where \( p_k \) is the normalized energy of the \( k \)-th frequency component. High entropy values indicate a complex, information-dense visual region, which is likely to be challenging for the model.

Using the spectral entropy scores, we identify the training images with the highest visual complexity as hard cases. Specifically, only those images with entropy values exceeding a predefined threshold \( H_{\text{min}} \) are selected. These high-entropy images are prioritized for additional training using the dual-learning loop described earlier, ensuring that the model focuses on challenging scenarios. By targeting complex visual patterns, the model improves its ability to handle interfaces with dense or intricate UI components, enhancing both referring and grounding performance.

\noindent\textbf{Functional Hard Case Mining.} Functional hard cases are identified based on the model's past performance. Specifically, we focus on samples where the model exhibits poor accuracy in interpreting functional descriptions. To enhance the model’s robustness in such cases, we use a description augmentation strategy that generates new, similar descriptions based on the problematic ones.

Let \( \mathcal{D}_{\text{hard}} \) represent the set of functional descriptions where the model has struggled. For each description \( D_i \in \mathcal{D}_{\text{hard}} \), we prompt a language model (e.g., GPT) to generate variations:

\begin{equation}
\text{Generate}(D_i) = \{D_i^{(1)}, D_i^{(2)}, \dots, D_i^{(n)}\}
\label{eq:generate}
\end{equation}

These augmented descriptions are then used as inputs to the dual-learning loop, creating synthetic functional hard cases. By iterating over these generated descriptions, the model incrementally improves its understanding of difficult functional concepts.

\noindent\textbf{Summary.} By leveraging the dual-learning loop and targeted hard case mining, \baby can autonomously discover and learn from difficult samples, thereby enhancing its robustness and adaptability in handling diverse GUI environments.

\section{Experiments}
\label{sec:experiments}
\begin{table*}[ht!]
    \centering
    \begin{tabular}{ccccccccc}
    \hline
    \multirow{2}{*}{\textbf{Model}} & \textbf{Number of} & \multicolumn{2}{c}{\textbf{Mobile}} & \multicolumn{2}{c}{\textbf{Desktop}} & \multicolumn{2}{c}{\textbf{Web}} & \multirow{2}{*}{\textbf{Avg.}} \\
    \cline{3-8}
     & \textbf{GUI Annotations} & \textbf{Text} & \textbf{Icon/Widget} & \textbf{Text} & \textbf{Icon/Widget} & \textbf{Text} & \textbf{Icon/Widget} &  \\
    \hline
    MiniGPT-v2\cite{chen2023minigpt} & -       & 8.4  & 6.6  & 6.2  & 2.9  & 6.5  & 3.4  & 5.7 \\ 
    GPT-4V\cite{achiam2023gpt4}     & -       & 22.6 & 24.5 & 20.2 & 11.8 & 9.2 & 8.8  & 16.2 \\
    GPT-4o\cite{achiam2023gpt4}     & -       & 20.2  & 24.9  & 21.1 & 23.6 & 12.2  & 7.8  & 18.3 \\
    Fuyu\cite{bavishi2023fuyu}      & -       & 41.0 & 1.3  & 33.0 & 3.6  & 33.9 & 4.4  & 19.5 \\
    CogAgent\cite{hong2024cogagent}   & 140M       & 67.0 & 24.0 & 74.2 & 20.0 & 70.4 & 28.6 & 47.4 \\
    SeeClick\cite{cheng2024seeclick}   & 850K       & 78.0 & 52.0 & 72.2 & 30.0 & 55.7 & 32.5 & 53.4 \\
    UGround\cite{gou2024navigating} & 10M       & 82.8 & 60.3 & 82.5 & \textbf{63.6} & 80.4 & 70.4 & 73.3 \\
    \hline
    \baby & 850K & \textbf{85.3} & \textbf{64.2} & \textbf{86.7} & 57.5 & \textbf{82.6} & \textbf{71.2} & \textbf{74.6} \\ 
    \hline
    \end{tabular}
    \vspace{-0.5em}
    \caption{Performance comparison of grounding models across different platforms and categories on the ScreenSpot benchmark.}
    \vspace{-0.5em}
    \label{tab:grounding_models_comparison}
\end{table*}

We present comprehensive experimental results to evaluate \baby's effectiveness across different aspects. Our evaluation focuses on three main areas: (1) GUI grounding capability on specialized benchmarks (\cref{sec:grounding_benchmarks}), demonstrating \baby's enhanced visual perception and functional understanding; (2) agent performance on real-world tasks (\cref{sec:agent_benchmarks}), showing \baby's practical utility in diverse applications; (3) ablation studies (\cref{sec:ablation}) examining the contribution of each component.

\subsection{Training Details}
\label{sec:training}
We follow the same training process as SeeClick\cite{cheng2024seeclick}, initializing from Qwen-VL\cite{bai2023qwenvl} with identical pretrained datasets: 850K GUI-specific data and 150K general vision-language instructions from the LLaVA\cite{liu2024llava} dataset. However, we incorporate two key enhancements:
\begin{itemize}
\item \textbf{Enhanced Visual Training}: We implement Information-Sensitive Cropping (ISC) during the initial training phase to improve visual perception capabilities.
\item \textbf{Self-Refining Stage}: Following the initial training, we conduct an additional SRDL stage where \baby is trained on approximately 3M self-annotated GUI samples generated through our dual-learning process.
\end{itemize}
Detailed hyperparameters and training configurations are provided in Appendix \cref{sup:training}.

\subsection{Evaluation on GUI Grounding Benchmarks}
\label{sec:grounding_benchmarks}
We evaluate \baby on two specialized GUI grounding benchmarks: ScreenSpot \cite{cheng2024seeclick} and GroundUI \cite{zheng2024agentstudio}, which test the model's ability to understand and interact with diverse interface elements across mobile, desktop, and web platforms. Details of the benchmarks and comparison models are provided in Appendix \cref{sup:benchmarks}.

On the ScreenSpot benchmark, \baby achieves state-of-the-art performance with an average accuracy of 74.6\%, outperforming previous models by significant margins (\cref{tab:grounding_models_comparison}). Detailed analysis reveals several key findings:

\begin{itemize}
    \item \textbf{Consistent Improvement over SeeClick}: Despite using the same amount of annotated training data (850K), \baby consistently outperforms SeeClick across all evaluated metrics, demonstrating the effectiveness of our ISC and SRDL enhancements in improving both visual perception and functional understanding capabilities.
    
    \item \textbf{Resolution-Dependent Performance Gain}: The performance improvements are more pronounced on web and desktop platforms compared to mobile. This pattern can be attributed to the inherent challenges of higher-resolution interfaces: web and desktop screens typically contain more complex layouts and denser information. Traditional approaches like SeeClick, which rely on uniform image scaling, inevitably lose fine-grained details in these complex interfaces. In contrast, our ISC method adaptively processes information-rich regions, effectively preserving critical visual details in high-resolution interfaces while maintaining computational efficiency.
    
    \item \textbf{Competitive Performance with Limited Annotations}: Compared to UGround, which utilizes more than 10 times our training data (10M vs 850K), \baby achieves superior performance in most categories and comparable overall accuracy. This demonstrates SRDL's effectiveness in identifying and learning from challenging cases, matching the performance of models trained on much larger annotated datasets.
\end{itemize}

\begin{table}[ht!]
    \centering
    \begin{tabular}{ccccc}
    \hline
    \textbf{Model} & \textbf{Web} & \textbf{Desktop} & \textbf{Mobile} & \textbf{Total} \\
    \hline
    Qwen-VL-Chat\cite{bai2023qwenvl} & 0.0 & 0.0 & 0.0 & 0.0 \\
    PaliGemma\cite{beyer2024paligemma} & 0.0 & 0.0 & 0.0 & 0.0 \\
    MiniCPM\cite{hu2024minicpm} & 0.0 & 0.3 & 2.7 & 0.9 \\
    Gemini-1.0\cite{team2023gemini} & 0.5 & 0.3 & 5.0 & 1.8 \\
    CogVLM2\cite{hong2024cogvlm2} & 2.5 & 2.7 & 5.3 & 3.4 \\
    GPT-4\cite{achiam2023gpt4} & 5.3 & 11.0 & 23.0 & 12.3 \\
    GPT-4o\cite{achiam2023gpt4} & 7.5 & 8.3 & 26.3 & 13.4 \\
    Claude\cite{anthropic2024claude} & 9.5 & 12.7 & 29.0 & 16.3 \\
    CogAgent\cite{hong2024cogagent} & 25.3 & 15.7 & 35.7 & 25.5 \\
    Gemini-1.5\cite{team2023gemini} & 31.2 & 24.3 & 51.3 & 35.2 \\
    SeeClick\cite{cheng2024seeclick} & 64.3 & 44.3 & 73.7 & 61.1 \\
    \hline
    \baby & \textbf{72.2} & \textbf{61.3} & \textbf{80.2} & \textbf{71.3} \\ %
    \hline
    \end{tabular}
    \vspace{-0.5em}
    \caption{Performance comparison of models across different platforms and total score on the GroundUI-1K benchmark.}
    \vspace{-0.5em}
    \label{tab:model_performance_comparison}
\end{table}

Similarly, on the GroundUI benchmark, \baby achieves \textit{sota} performance across all categories, outperforming previous models by significant margins (\cref{tab:model_performance_comparison}).

These results validate our key technical contributions: (1) ISC effectively enhances the model's visual perception, particularly for high-resolution interfaces, and (2) SRDL successfully mines and learns from challenging cases, achieving performance comparable to or better than models trained with 10x more annotated data.

\subsection{Evaluation on Agent Benchmarks}
\label{sec:agent_benchmarks}

\begin{table*}[t!]
    \centering
    \begin{tabular}{ccccccccccc}
    \hline
    \multirow{2}{*}{\textbf{Model}} & \textbf{w/o} & \multicolumn{3}{c}{\textbf{Cross-Task}} & \multicolumn{3}{c}{\textbf{Cross-Website}} & \multicolumn{3}{c}{\textbf{Cross-Domain}} \\
    \cline{3-11}
     & \textbf{HTML} & \textbf{Ele.Acc} & \textbf{Op.F1} & \textbf{Step SR} & \textbf{Ele.Acc} & \textbf{Op.F1} & \textbf{Step SR} & \textbf{Ele.Acc} & \textbf{Op.F1} & \textbf{Step SR} \\
    \hline
    MindAct (gen)\cite{deng2024mind2web} & ✗ & 20.2 & 52.0 & 17.5 & 13.9 & 44.7 & 11.0 & 14.2 & 44.7 & 11.9 \\
    MindAct\cite{deng2024mind2web} & ✗ & \textbf{55.1} & 75.7 & \textbf{52.0} & \textbf{42.0} & 65.2 & \textbf{38.9} & \textbf{42.1} & 66.5 & \textbf{39.6} \\
    GPT-3.5-Turbo\cite{achiam2023gpt4} & ✗ & 20.3 & 56.6 & 17.4 & 19.3 & 48.8 & 16.2 & 21.6 & 52.8 & 18.6 \\
    GPT-4\cite{achiam2023gpt4} & ✗ & 41.6 & 60.6 & \underline{36.2} & 35.8 & 51.1 & 30.1 & 37.1 & 46.5 & 26.4 \\
    Qwen-VL\cite{bai2023qwenvl} & ✓ & 15.9 & 86.7 & 13.3 & 13.2 & \textbf{83.5} & 9.2 & 14.1 & 84.3 & 12.0 \\
    SeeClick\cite{cheng2024seeclick} & ✓ & 28.3 & \underline{87.0} & 25.5 & 21.4 & 80.6 & 16.4 & 23.2 & \underline{84.8} & 20.8 \\
    \hline
    \baby & ✓ & \underline{33.5} & \textbf{87.1} & 32.0 & \underline{31.2} & \underline{82.2} & \underline{26.2} & \underline{32.8} & \textbf{85.1} & \underline{28.8} \\ %
    \hline
    \end{tabular}
    \vspace{-0.5em}
    \caption{Performance comparison on Mind2Web benchmark across different evaluation settings. For each metric, the \textbf{bold} values indicate best performance and \underline{underlined} values indicate second best.}
    \vspace{-0.5em}
    \label{tab:methods_performance_comparison}
\end{table*}

\begin{table*}[h!]
    \centering
    \begin{tabular}{ccccccccc}
    \hline
    \textbf{Methods} & \textbf{Modality} & \textbf{General} & \textbf{Install} & \textbf{GoogleApps} & \textbf{Single} & \textbf{WebShopping} & \textbf{Overall} & \textbf{ClickAcc} \\
    \hline
    ChatGPT-CoT\cite{zhang2023you} & Text & 5.9 & 4.4 & 10.5 & 9.4 & 8.4 & 7.7 & - \\
    PaLM2-CoT\cite{rawles2024androidinthewild} & Text & - & - & - & - & - & 39.6 & - \\
    GPT-4V\cite{achiam2023gpt4} & Image & 41.7 & 42.6 & 49.8 & \textbf{72.8} & 45.7 & 50.5 & - \\
    Qwen-VL\cite{bai2023qwenvl} & Image & 49.5 & 59.9 & 46.9 & 64.7 & 50.7 & 54.3 & 57.4 \\
    SeeClick\cite{cheng2024seeclick} & Image & 54.0 & 66.4 & 54.9 & 63.5 & 57.6 & 59.3 & 66.4 \\
    \hline
    \baby & Image & \textbf{61.5} & \textbf{71.4} & \textbf{58.3} & 66.4 & \textbf{60.2} & \textbf{63.6} & \textbf{71.0} \\ 
    \hline
    \end{tabular}
    \vspace{-0.5em}
    \caption{Average scores of different methods on AITW. The best results in each column are \textbf{bold}.}
    \vspace{-0.5em}
    \label{tab:average_scores_aitw}
\end{table*}

We assess \baby's practical utility on two comprehensive agent benchmarks: Mind2Web\cite{deng2024mind2web} for web interactions and Android In the Wild (AITW)\cite{rawles2024androidinthewild} for mobile OS tasks (\cref{tab:methods_performance_comparison} and~\cref{tab:average_scores_aitw}). We follow the same evaluation setting as SeeClick \cite{cheng2024seeclick}. The detail of the benchmarks and comparison models are provided in Appendix \cref{sup:agent}.

Across both benchmarks, \baby consistently outperforms SeeClick by significant margins, achieving best performance in 11/12 categories, among pure GUI-based models (w/o HTML). The gains are particularly notable in complex scenarios requiring precise element localization and multi-step interactions, where \baby's superior understanding of both visual layouts and functional relationships proves especially beneficial. This consistent improvement across diverse agent tasks validates our approach of strengthening foundational GUI understanding through ISC and SRDL, showing that better visual perception and more robust grounding capabilities are crucial for advancing real-world agent applications.

\subsection{Ablation Study}
\label{sec:ablation}

\noindent\textbf{Overall Component Analysis: ISC and SRDL are Complementary.} We conduct comprehensive ablation studies to analyze the effectiveness of our two key components: Information-Sensitive Cropping (ISC) and Self-Refining Dual Learning (SRDL). As shown in the figure, both components contribute uniquely to \baby's performance. The full \baby model achieves the optimal balance, reaching approximately 75\% accuracy at 1.0s processing time. This represents a significant improvement over baseline methods like SeeClick (~53\% accuracy at 0.5s) and CogAgent (~48\% accuracy at 2.0s). ISC provides a 300\% efficiency improvement, while SRDL contributes a 10\% accuracy gain. The complementary nature of these components allows \baby to achieve performance comparable to UGround (shown at 3.0-3.5s) but with significantly faster processing time.

\begin{figure}[h!]
    \centering
    \includegraphics[width=\columnwidth]{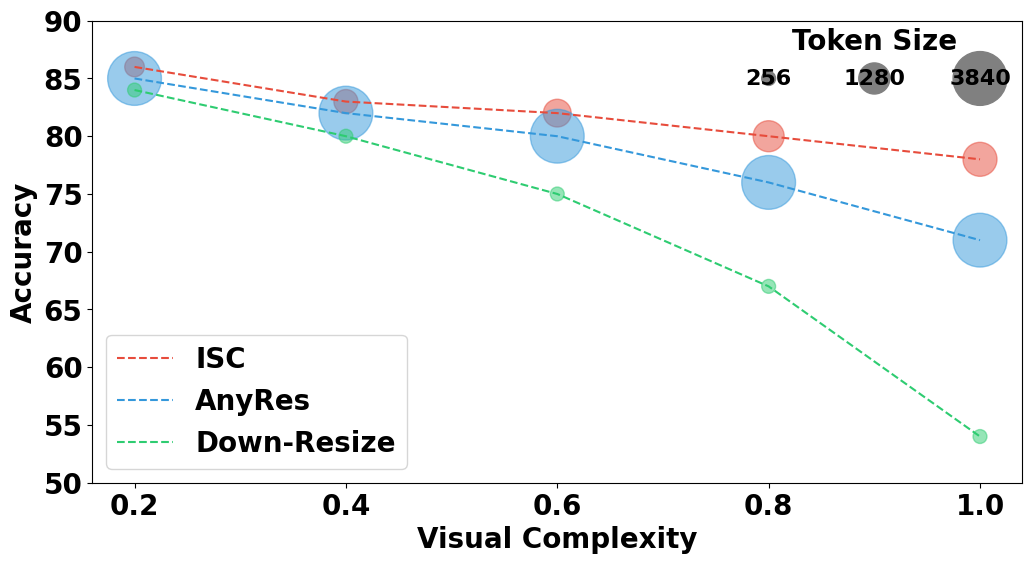}
    \vspace{-0.5em}
    \caption{ISC's effectiveness across different visual complexity.}
    \label{fig:ablation_isc}
    \vspace{-0.5em}
\end{figure}

\noindent\textbf{ISC Analysis: Optimizing Token Allocation Efficiency Across Visual Complexities.} Our analysis reveals ISC's superior efficiency in token utilization when compared to baseline approaches (down-resize and AnyRes adaptive processing). As illustrated in \cref{fig:ablation_isc}, where bubble sizes indicate token count, ISC demonstrates intelligent resource allocation by using minimal tokens for simple interfaces while dynamically scaling up only when needed for complex scenes. At low visual complexity, ISC achieves high accuracy with significantly fewer tokens than baselines. When visual complexity increases, ISC automatically allocates more tokens to capture more detailed information, balancing accuracy and efficiency. The results show that ISC's key strength lies in its ability to adapt token allocation based on information density, achieving better accuracy than baseline approaches while consistently using fewer computational resources across all complexity levels.

\begin{figure}[h!]
    \centering
    \includegraphics[width=\columnwidth]{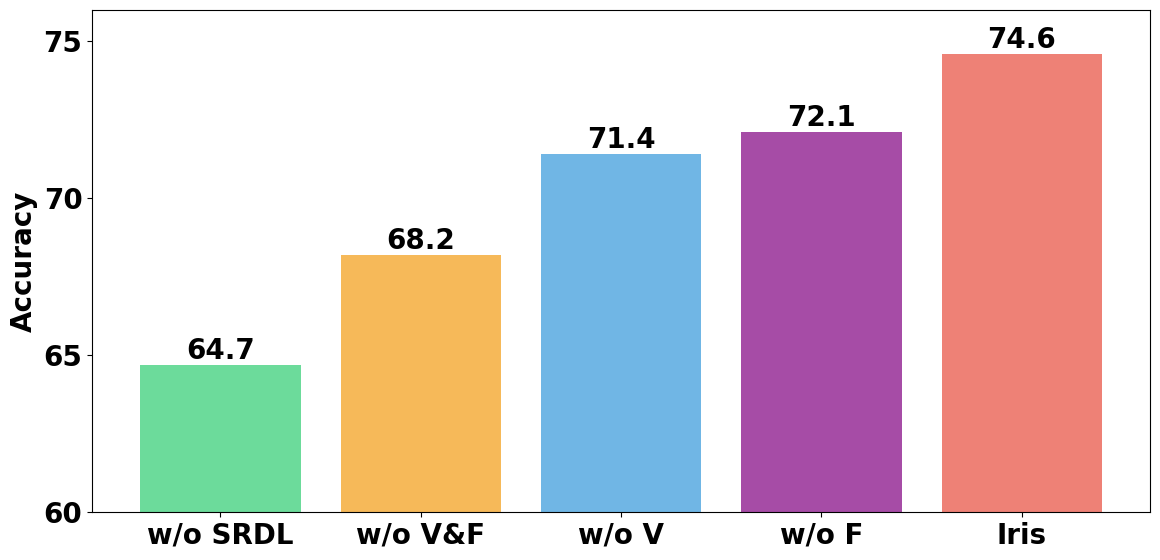}
    \vspace{-0.5em}
    \caption{SRDL's effectiveness across different case mining strategies.}
    \label{fig:ablation_srdl}
    \vspace{-0.5em}
\end{figure}

\noindent\textbf{SRDL Analysis: Both Visual and Functional Hard Case Mining are Essential.} Our analysis of SRDL's hard case mining strategy reveals the critical role of both visual and functional components. As shown in \cref{fig:ablation_srdl}, the baseline model without SRDL achieves 64.7\% accuracy, while incorporating SRDL with either visual or functional mining disabled (w/o V and w/o F) yields improved but still suboptimal results of 71.4\% and 72.1\% respectively. The full SRDL implementation, combining both mining strategies, achieves the highest accuracy of 74.6\%. This demonstrates how visual mining, which identifies challenging elements based on appearance and layout, complements functional mining, which focuses on complex interaction patterns. Together, they create a comprehensive system for improving model robustness across different types of challenging cases.
\section{Conclusion}
\label{sec:conclusion}

This paper introduces \baby, a foundational visual agent that advances GUI understanding through two key innovations: Information-Sensitive Cropping (ISC) and Self-Refining Dual Learning (SRDL). ISC enables efficient processing of high-resolution interfaces by adaptively allocating computational resources based on information density, achieving a 300\% efficiency improvement over traditional approaches. Rather than treating all regions equally, ISC employs edge detection to identify visually dense areas and dynamically adjusts token allocation, allowing for detailed processing of complex UI elements while efficiently handling sparse regions. SRDL enhances model robustness through autonomous discovery and learning from challenging cases, delivering a 10\% accuracy gain without requiring additional manual annotations. Evaluations on multiple benchmarks, ranging from basic GUI grounding to complex multimodal agent tasks, demonstrate that \baby achieves \textit{state-of-the-art} performance while balancing both accuracy and efficiency, setting a new standard for GUI-based digital agents.

\clearpage

\clearpage
\setcounter{page}{1}
\maketitlesupplementary

\section{ISC Implementation Details}\label{sup:isc}

\subsection{Motivation}
Information detection in GUI interfaces presents unique challenges due to the coexistence of information-dense regions (e.g., menu bars, toolbars) and sparse areas (e.g., backgrounds, margins). Our implementation aims to efficiently identify these varying density regions while maintaining real-time processing capabilities.

\subsection{Technical Approach}
Our information detection approach builds on three key insights:

\begin{enumerate}
    \item \textbf{Edge Significance:} GUI elements typically have distinct boundaries that separate them from backgrounds and other elements. By detecting these edges, we can identify regions containing meaningful interface components.
    
    \item \textbf{Adaptive Processing:} Rather than applying uniform processing across the entire interface, we use adaptive histogram equalization to enhance local contrast. This helps identify subtle UI elements while maintaining sensitivity to prominent features.
    
    \item \textbf{Density Preservation:} The detected edges are deliberately dilated to create connected regions that better represent the actual information density of UI components. This prevents fragmentation of logical interface elements and helps maintain semantic grouping.
\end{enumerate}

\subsection{Edge Detection Pipeline}
The edge detection \cite{canny1986computational} process consists of four sequential stages:

\begin{enumerate}
    \item \textbf{Pre-processing:}
    \begin{itemize}
        \item Convert the input image to grayscale to focus on structural information
        \item Apply adaptive histogram equalization to enhance local contrasts
        \item This stage ensures that both prominent and subtle interface elements are captured
    \end{itemize}

    \item \textbf{Noise Reduction:}
    \begin{itemize}
        \item Apply Gaussian smoothing to reduce image noise
        \item This step prevents false edge detection from texture and compression artifacts
        \item The smoothing preserves significant boundaries while eliminating minor variations
    \end{itemize}

    \item \textbf{Gradient Computation:}
    \begin{itemize}
        \item Calculate intensity gradients in both horizontal and vertical directions
        \item Compute gradient magnitude and direction at each pixel
        \item Strong gradients indicate potential edges in the interface
    \end{itemize}

    \item \textbf{Edge Formation:}
    \begin{itemize}
        \item Apply non-maximum suppression to thin edge responses
        \item Use hysteresis thresholding to connect edge segments
        \item The result is a binary edge map highlighting significant UI boundaries
    \end{itemize}
\end{enumerate}

\subsection{Working Mechanism}
The detection process operates by:
\begin{enumerate}
    \item Converting the input GUI screenshot to a representation that emphasizes structural information
    \item Enhancing local contrasts to identify both prominent and subtle interface elements
    \item Detecting and strengthening edges to create a cohesive information density map
    \item Producing a binary matrix that indicates regions of high information content
\end{enumerate}

The resulting information matrix serves as a guide for the Information-Sensitive Cropping (ISC) process described in the main paper, enabling efficient allocation of computational resources based on the actual distribution of interface elements.

\section{ISC Efficiency Analysis}\label{sup:isc_efficiency}

\subsection{ISC Computational Complexity}

The ISC pipeline consists of two main stages:

\textbf{1. Information Detection}

Edge detection has a complexity of $O(W \times H)$, where $W$ and $H$ are the image width and height. This includes:
\begin{itemize}
    \item Grayscale conversion, adaptive histogram equalization, and edge detection operations
    \item Each operation has linear complexity in terms of image pixels
\end{itemize}

\textbf{2. Adaptive Cropping}

For sliding windows of size $k$:
\begin{itemize}
    \item Step size is $k/4$
    \item Density calculation per position: $O(k^2)$
    \item Number of windows: $O((W \times H)/(k/4)^2)$
    \item Window sizes increase geometrically from $k_{min}$ to $max(W,H)$
\end{itemize}
Total complexity: $O(W \times H)$

\subsection{MLLM Inference Complexity Comparison}

Let $H$ be the number of attention heads and $d$ be the dimension of each head in the MLLM. The computational complexity can be analyzed as follows:

\noindent\textbf{1. Without ISC}

For standard MLLM processing with direct input of full resolution images:
\begin{itemize}
    \item Input image size: $W \times H$
    \item Visual token count: $N_{full} = \lceil\frac{W \times H}{p^2}\rceil$, where $p$ is the patch size
    \item Each self-attention operation: $O(N_{full}^2 \cdot d)$
    \item Across all heads: $O(N_{full}^2 \cdot H \cdot d)$
    \item Total complexity: $T_{standard} = O((W \times H / p^2)^2 \cdot H \cdot d)$
\end{itemize}

\noindent\textbf{2. With ISC}

Using ISC-based processing with adaptive sub-image selection:
\begin{itemize}
    \item ISC preprocessing: $O(W \times H)$
    \item Maximum sub-images: $M$ (fixed constant)
    \item Tokens per sub-image: $N_{sub} = (S/p)^2$, where $S$ is the target size
    \item Attention complexity per sub-image: $O(N_{sub}^2 \cdot H \cdot d)$
    \item Total complexity: $T_{ISC} = O(W \times H + M \cdot N_{sub}^2 \cdot H \cdot d)$
\end{itemize}

Note that $M$ and $S$ can be flexibly adjusted according to computational budget and model performance requirements. The key efficiency gain comes from the fact that $T_{ISC}$ scales linearly with input resolution $(W \times H)$, while $T_{standard}$ scales quadratically. This difference becomes particularly significant for high-resolution GUI interfaces where $(W \times H) \gg (M \cdot S^2)$.

\subsection{ISC Efficiency Benefits}
This results in significant efficiency improvements:
\begin{itemize}
    \item ISC reduces quadratic dependency on input resolution to linear
    \item Total computation becomes bounded by a constant factor of sub-image count
    \item The theoretical speedup ratio $\frac{T_{standard}}{T_{ISC}}$ increases with input resolution
    \item For typical GUI resolutions (1920×1080), ISC achieves approximately 300\% speedup in practice
\end{itemize}

\section{\baby Training Details}\label{sup:training}
Following SeeClick \cite{cheng2024seeclick}, we employed the same training data (850K GUI-specific data and 150K general vision-language instructions from the LLaVA \cite{liu2024llava} dataset) for continual pre-training of Qwen-VL-Chat \cite{bai2023qwenvl}, only with a different image processing pipeline, where SeeClick resizes the input image to a fixed resolution of 448x448 but we use our ISC pipeline to select the most relevant sub-images. Then we had another Self-Refining Dual Learning stage to further improve the model's understanding of complex GUI interfaces. Based on the training history of the first stage, we further self-labeled another 3M data, as introduced in \ref{sec:srdl}. Then we had another round of training with the new data mixed with the original training data, with all other training hyper-parameters kept the same. The optimizer we used is AdamW \cite{kingma2014adam} and we picked a cosine annealing scheduler with an init learning rate of 3e-5 and a global batch size of 64.

\section{GUI Grounding Benchmarks Details}\label{sup:benchmarks}

\subsection{SeeClick Benchmark}
We evaluate our model on ScreenSpot, a comprehensive benchmark proposed by Cheng et al. \cite{cheng2024seeclick}. ScreenSpot contains 610 interface screenshots with 1,272 annotated instructions spanning various platforms (iOS, Android, macOS, Windows, and web). The instruction set is divided into 502 mobile, 334 desktop and 436 web instructions. Each instruction is paired with precise bounding box coordinates marking the target UI element.

The data collection process of ScreenSpot follows rigorous protocols where:
\begin{itemize}
   \item Screenshots are captured during daily use by experienced annotators
   \item Instructions are written in natural language describing common operations
   \item Both text elements and UI widgets/icons are annotated
   \item Platform distribution is balanced to cover:
   \begin{itemize}
       \item Mobile: iOS and Android mobile interfaces
       \item Desktop: macOS and Windows operating systems  
       \item Web: Development, shopping, forum and tool websites
   \end{itemize}
\end{itemize}

\subsection{GroundUI Benchmark}
For additional validation, we utilize the GroundUI benchmark from AgentStudio \cite{zheng2024agentstudio} which curates data from multiple existing datasets:

\textbf{Dataset Composition:}

Each sample consists of:
\begin{itemize}
   \item An interface screenshot
   \item A single-step natural language instruction
   \item Ground truth bounding box coordinates
\end{itemize}

For efficient evaluation, we use GroundUI-1K, a carefully selected subset that maintains the diversity of the full dataset:
\begin{itemize}
   \item 400 web samples covering various website types:
   \begin{itemize}
       \item Development tools and environments
       \item E-commerce and shopping platforms
       \item Forum and community websites
       \item Utility and productivity tools
   \end{itemize}
   \item 300 desktop samples from:
   \begin{itemize}
       \item System settings and configurations
       \item Common productivity applications
       \item Built-in utilities
   \end{itemize}
   \item 300 mobile samples featuring:
   \begin{itemize}
       \item Native system applications
       \item Third-party mobile apps
       \item Mobile web browsers
   \end{itemize}
\end{itemize}

\subsection{Compared Baselines}
Following previous works, we compare our approach with several state-of-the-art vision-language models:

\textbf{Open-source Models:}
\begin{itemize}
   \item PaliGemma \cite{beyer2024paligemma}: A 3B parameter versatile vision-language model with flexible resolution support up to 896x896 
   \item MiniCPM \cite{hu2024minicpm}: A small yet efficient multimodal model built with scalable training strategies
   \item CogVLM2 \cite{hong2024cogvlm2}: An improved vision-language model with enhanced image and video understanding capabilities 
   \item Qwen-VL-Chat \cite{bai2023qwenvl}: A vision-language model with open-ended dialogue abilities 
   \item MiniGPT-v2 \cite{chen2023minigpt}: A 7B parameter model for multi-task vision-language learning 
\end{itemize}

\textbf{Proprietary Models:}
\begin{itemize}
   \item Gemini-1.0/1.5 \cite{team2023gemini}: Google's multimodal models with progressive improvements in visual understanding and reasoning 
   \item GPT-4o/GPT-4V \cite{achiam2023gpt4}: OpenAI's large language model with strong zero-shot visual capabilities.  
   \item Claude \cite{anthropic2024claude}: Anthropic's multimodal model demonstrating strong performance in language tasks with visual inputs 
\end{itemize}

\textbf{GUI-Specialized Models:}
\begin{itemize}
   \item Fuyu \cite{bavishi2023fuyu}: A model architecture which takes naive image as input
   \item CogAgent \cite{hong2024cogagent}: A 140M-sample GUI-pretrained model focused on computer interaction tasks 
   \item UGround \cite{gou2024navigating}: A universal visual grounding model trained on 10M GUI samples 
   \item SeeClick \cite{cheng2024seeclick}: A GUI-focused model pretrained on 850K samples with specialized grounding capabilities 
\end{itemize}

For fair comparison, we evaluate all models using the same click accuracy metric - considering a prediction successful only when the predicted coordinates fall within the ground truth element's bounding box. This directly measures each model's ability to precisely locate and interact with UI elements based on natural language instructions.

\section{Agent Benchmark Details}\label{sup:agent}

We evaluate Iris on two comprehensive agent benchmarks following the same settings as SeeClick \cite{cheng2024seeclick}: AITW \cite{rawles2024androidinthewild} for mobile OS interactions and Mind2Web \cite{deng2024mind2web} for web navigation.

\subsection{Android In The Wild (AITW)}
AITW \cite{rawles2024androidinthewild} is a comprehensive Android automation dataset containing over 30K instructions and 700K episodes across five categories: General, Install, GoogleApps, Single, and WebShopping. Each sample contains an instruction and corresponding action trajectory with screenshots.

For evaluation, we follow SeeClick's approach in using a screen-wise matching score that considers both the correctness of action type and its value (e.g., click coordinates or typed text). This scoring mechanism has been shown to correlate well with human-judged task completion rates. We utilize their standard metrics while ensuring fair comparisons across different models.

\subsection{Mind2Web}
Mind2Web \cite{deng2024mind2web} is a web automation benchmark involving real-world websites, consisting of over 2000 open-ended tasks collected from 137 different websites. Each task includes high-level instructions and corresponding human action trajectories.

Following SeeClick, we evaluate using three key metrics:

\begin{itemize}
    \item Element Accuracy (Ele.Acc): For visual-based methods like Iris, we consider a prediction correct if the predicted click coordinates fall within the ground truth element's bounding box
    
    \item Operation F1 (Op.F1): For click operations, this is equivalent to accuracy. For type and select operations, it calculates the token-level F1 score of predicted values
    
    \item Step Success Rate (Step SR): A step is considered successful only if both the predicted element and operation are correct
\end{itemize}

We maintain strict evaluation criteria where predictions falling outside the target element's bounding box are considered incorrect, even if they are near the target. This ensures consistency with previous evaluations while providing clear metrics for assessing localization accuracy.

\clearpage
{
    \small
    \bibliographystyle{ieeenat_fullname}
    \bibliography{main} 

\begin{thebibliography}{43}
\providecommand{\natexlab}[1]{#1}
\providecommand{\url}[1]{\texttt{#1}}
\expandafter\ifx\csname urlstyle\endcsname\relax
  \providecommand{\doi}[1]{doi: #1}\else
  \providecommand{\doi}{doi: \begingroup \urlstyle{rm}\Url}\fi

\bibitem[Achiam et~al.(2023{\natexlab{a}})Achiam, Adler, Agarwal, Ahmad, Akkaya, Aleman, Almeida, Altenschmidt, Altman, Anadkat, et~al.]{achiam2023gpt}
Josh Achiam, Steven Adler, Sandhini Agarwal, Lama Ahmad, Ilge Akkaya, Florencia~Leoni Aleman, Diogo Almeida, Janko Altenschmidt, Sam Altman, Shyamal Anadkat, et~al.
\newblock Gpt-4 technical report.
\newblock \emph{arXiv preprint arXiv:2303.08774}, 2023{\natexlab{a}}.

\bibitem[Achiam et~al.(2023{\natexlab{b}})Achiam, Adler, Agarwal, Ahmad, Akkaya, Aleman, Almeida, Altenschmidt, Altman, Anadkat, et~al.]{achiam2023gpt4}
Josh Achiam, Steven Adler, Sandhini Agarwal, Lama Ahmad, Ilge Akkaya, Florencia~Leoni Aleman, Diogo Almeida, Janko Altenschmidt, Sam Altman, Shyamal Anadkat, et~al.
\newblock Gpt-4 technical report.
\newblock \emph{arXiv preprint arXiv:2303.08774}, 2023{\natexlab{b}}.

\bibitem[Anthropic(2024)]{anthropic2024claude}
AI Anthropic.
\newblock The claude 3 model family: Opus, sonnet, haiku.
\newblock \emph{Claude-3 Model Card}, 1, 2024.

\bibitem[Bai et~al.(2023)Bai, Bai, Yang, Wang, Tan, Wang, Lin, Zhou, and Zhou]{bai2023qwenvl}
Jinze Bai, Shuai Bai, Shusheng Yang, Shijie Wang, Sinan Tan, Peng Wang, Junyang Lin, Chang Zhou, and Jingren Zhou.
\newblock Qwen-vl: A frontier large vision-language model with versatile abilities.
\newblock \emph{arXiv preprint arXiv:2308.12966}, 2023.

\bibitem[Bavishi et~al.(2023)Bavishi, Elsen, Hawthorne, Nye, Odena, Somani, and Ta{\c{s}}{\i}rlar]{bavishi2023fuyu}
Rohan Bavishi, Erich Elsen, Curtis Hawthorne, Maxwell Nye, Augustus Odena, Arushi Somani, and Sa{\u{g}}nak Ta{\c{s}}{\i}rlar.
\newblock Fuyu-8b: A multimodal architecture for ai agents, 2023.

\bibitem[Beyer et~al.(2024)Beyer, Steiner, Pinto, Kolesnikov, Wang, Salz, Neumann, Alabdulmohsin, Tschannen, Bugliarello, et~al.]{beyer2024paligemma}
Lucas Beyer, Andreas Steiner, Andr{\'e}~Susano Pinto, Alexander Kolesnikov, Xiao Wang, Daniel Salz, Maxim Neumann, Ibrahim Alabdulmohsin, Michael Tschannen, Emanuele Bugliarello, et~al.
\newblock Paligemma: A versatile 3b vlm for transfer.
\newblock \emph{arXiv preprint arXiv:2407.07726}, 2024.

\bibitem[Canny(1986)]{canny1986computational}
John Canny.
\newblock A computational approach to edge detection.
\newblock \emph{IEEE Transactions on pattern analysis and machine intelligence}, \penalty0 (6):\penalty0 679--698, 1986.

\bibitem[Chen et~al.(2023)Chen, Zhu, Shen, Li, Liu, Zhang, Krishnamoorthi, Chandra, Xiong, and Elhoseiny]{chen2023minigpt}
Jun Chen, Deyao Zhu, Xiaoqian Shen, Xiang Li, Zechun Liu, Pengchuan Zhang, Raghuraman Krishnamoorthi, Vikas Chandra, Yunyang Xiong, and Mohamed Elhoseiny.
\newblock Minigpt-v2: large language model as a unified interface for vision-language multi-task learning.
\newblock \emph{arXiv preprint arXiv:2310.09478}, 2023.

\bibitem[Cheng et~al.(2024)Cheng, Sun, Chu, Xu, Li, Zhang, and Wu]{cheng2024seeclick}
Kanzhi Cheng, Qiushi Sun, Yougang Chu, Fangzhi Xu, Yantao Li, Jianbing Zhang, and Zhiyong Wu.
\newblock Seeclick: Harnessing gui grounding for advanced visual gui agents.
\newblock \emph{arXiv preprint arXiv:2401.10935}, 2024.

\bibitem[Deka et~al.(2017)Deka, Huang, Franzen, Hibschman, Afergan, Li, Nichols, and Kumar]{deka2017rico}
Biplab Deka, Zifeng Huang, Chad Franzen, Joshua Hibschman, Daniel Afergan, Yang Li, Jeffrey Nichols, and Ranjitha Kumar.
\newblock Rico: A mobile app dataset for building data-driven design applications.
\newblock In \emph{Proceedings of the 30th annual ACM symposium on user interface software and technology}, pages 845--854, 2017.

\bibitem[Deng et~al.(2024)Deng, Gu, Zheng, Chen, Stevens, Wang, Sun, and Su]{deng2024mind2web}
Xiang Deng, Yu Gu, Boyuan Zheng, Shijie Chen, Sam Stevens, Boshi Wang, Huan Sun, and Yu Su.
\newblock Mind2web: Towards a generalist agent for the web.
\newblock \emph{Advances in Neural Information Processing Systems}, 36, 2024.

\bibitem[Dong et~al.(2024{\natexlab{a}})Dong, Zhang, Zang, Cao, Wang, Ouyang, Zhang, Duan, Zhang, Li, et~al.]{dong2024internlm}
Xiaoyi Dong, Pan Zhang, Yuhang Zang, Yuhang Cao, Bin Wang, Linke Ouyang, Songyang Zhang, Haodong Duan, Wenwei Zhang, Yining Li, et~al.
\newblock Internlm-xcomposer2-4khd: A pioneering large vision-language model handling resolutions from 336 pixels to 4k hd.
\newblock \emph{arXiv preprint arXiv:2404.06512}, 2024{\natexlab{a}}.

\bibitem[Dong et~al.(2024{\natexlab{b}})Dong, Zhang, Zang, Cao, Wang, Ouyang, Zhang, Duan, Zhang, Li, et~al.]{dong2024internlm4k}
Xiaoyi Dong, Pan Zhang, Yuhang Zang, Yuhang Cao, Bin Wang, Linke Ouyang, Songyang Zhang, Haodong Duan, Wenwei Zhang, Yining Li, et~al.
\newblock Internlm-xcomposer2-4khd: A pioneering large vision-language model handling resolutions from 336 pixels to 4k hd.
\newblock \emph{arXiv preprint arXiv:2404.06512}, 2024{\natexlab{b}}.

\bibitem[Fei et~al.(2024{\natexlab{a}})Fei, Wu, Zhang, Chua, and Yan]{fei2024vitron}
Hao Fei, Shengqiong Wu, Hanwang Zhang, Tat-Seng Chua, and Shuicheng Yan.
\newblock Vitron: A unified pixel-level vision llm for understanding, generating, segmenting, editing.
\newblock 2024{\natexlab{a}}.

\bibitem[Fei et~al.(2024{\natexlab{b}})Fei, Wu, Zhang, Zhang, Chua, and Yan]{fei2024enhancing}
Hao Fei, Shengqiong Wu, Meishan Zhang, Min Zhang, Tat-Seng Chua, and Shuicheng Yan.
\newblock Enhancing video-language representations with structural spatio-temporal alignment.
\newblock \emph{IEEE Transactions on Pattern Analysis and Machine Intelligence}, 2024{\natexlab{b}}.

\bibitem[Gao et~al.(2024{\natexlab{a}})Gao, Bu, Miao, Wu, Li, Li, Tang, Wu, Zhuang, and Wang]{gao2024generalist}
Minghe Gao, Wendong Bu, Bingchen Miao, Yang Wu, Yunfei Li, Juncheng Li, Siliang Tang, Qi Wu, Yueting Zhuang, and Meng Wang.
\newblock Generalist virtual agents: A survey on autonomous agents across digital platforms.
\newblock \emph{arXiv preprint arXiv:2411.10943}, 2024{\natexlab{a}}.

\bibitem[Gao et~al.(2024{\natexlab{b}})Gao, Li, Fei, Pang, Ji, Wang, Lv, Zhang, Tang, and Zhuang]{gao2024fine}
Minghe Gao, Juncheng Li, Hao Fei, Liang Pang, Wei Ji, Guoming Wang, Zheqi Lv, Wenqiao Zhang, Siliang Tang, and Yueting Zhuang.
\newblock De-fine: De composing and re fin ing visual programs with auto-feedback.
\newblock In \emph{Proceedings of the 32nd ACM International Conference on Multimedia}, pages 7649--7657, 2024{\natexlab{b}}.

\bibitem[Gou et~al.(2024)Gou, Wang, Zheng, Xie, Chang, Shu, Sun, and Su]{gou2024navigating}
Boyu Gou, Ruohan Wang, Boyuan Zheng, Yanan Xie, Cheng Chang, Yiheng Shu, Huan Sun, and Yu Su.
\newblock Navigating the digital world as humans do: Universal visual grounding for gui agents.
\newblock \emph{arXiv preprint arXiv:2410.05243}, 2024.

\bibitem[Hong et~al.(2024{\natexlab{a}})Hong, Wang, Ding, Yu, Lv, Wang, Cheng, Huang, Ji, Xue, et~al.]{hong2024cogvlm2}
Wenyi Hong, Weihan Wang, Ming Ding, Wenmeng Yu, Qingsong Lv, Yan Wang, Yean Cheng, Shiyu Huang, Junhui Ji, Zhao Xue, et~al.
\newblock Cogvlm2: Visual language models for image and video understanding.
\newblock \emph{arXiv preprint arXiv:2408.16500}, 2024{\natexlab{a}}.

\bibitem[Hong et~al.(2024{\natexlab{b}})Hong, Wang, Lv, Xu, Yu, Ji, Wang, Wang, Dong, Ding, et~al.]{hong2024cogagent}
Wenyi Hong, Weihan Wang, Qingsong Lv, Jiazheng Xu, Wenmeng Yu, Junhui Ji, Yan Wang, Zihan Wang, Yuxiao Dong, Ming Ding, et~al.
\newblock Cogagent: A visual language model for gui agents.
\newblock In \emph{Proceedings of the IEEE/CVF Conference on Computer Vision and Pattern Recognition}, pages 14281--14290, 2024{\natexlab{b}}.

\bibitem[Hu et~al.(2024)Hu, Tu, Han, He, Cui, Long, Zheng, Fang, Huang, Zhao, et~al.]{hu2024minicpm}
Shengding Hu, Yuge Tu, Xu Han, Chaoqun He, Ganqu Cui, Xiang Long, Zhi Zheng, Yewei Fang, Yuxiang Huang, Weilin Zhao, et~al.
\newblock Minicpm: Unveiling the potential of small language models with scalable training strategies.
\newblock \emph{arXiv preprint arXiv:2404.06395}, 2024.

\bibitem[Kingma(2014)]{kingma2014adam}
Diederik~P Kingma.
\newblock Adam: A method for stochastic optimization.
\newblock \emph{arXiv preprint arXiv:1412.6980}, 2014.

\bibitem[Li et~al.(2023)Li, Pan, Ge, Gao, Zhang, Ji, Zhang, Chua, Tang, and Zhuang]{li2023finetuning}
Juncheng Li, Kaihang Pan, Zhiqi Ge, Minghe Gao, Hanwang Zhang, Wei Ji, Wenqiao Zhang, Tat-Seng Chua, Siliang Tang, and Yueting Zhuang.
\newblock Fine-tuning multimodal llms to follow zero-shot demonstrative instructions.
\newblock \emph{arXiv preprint arXiv:2308.04152}, 2023.

\bibitem[Li et~al.(2024)Li, You, Zhang, Feng, Agrawal, Li, Moorthy, Nichols, Yang, and Gan]{li2024ferret2}
Zhangheng Li, Keen You, Haotian Zhang, Di Feng, Harsh Agrawal, Xiujun Li, Mohana Prasad~Sathya Moorthy, Jeff Nichols, Yinfei Yang, and Zhe Gan.
\newblock Ferret-ui 2: Mastering universal user interface understanding across platforms.
\newblock \emph{arXiv preprint arXiv:2410.18967}, 2024.

\bibitem[Liu et~al.(2024{\natexlab{a}})Liu, Zhang, Qiu, Huang, Lin, Zhao, Geng, Lin, Jin, Zhang, et~al.]{liu2024sphinx}
Dongyang Liu, Renrui Zhang, Longtian Qiu, Siyuan Huang, Weifeng Lin, Shitian Zhao, Shijie Geng, Ziyi Lin, Peng Jin, Kaipeng Zhang, et~al.
\newblock Sphinx-x: Scaling data and parameters for a family of multi-modal large language models.
\newblock \emph{arXiv preprint arXiv:2402.05935}, 2024{\natexlab{a}}.

\bibitem[Liu et~al.(2024{\natexlab{b}})Liu, Li, Wu, and Lee]{liu2024llava}
Haotian Liu, Chunyuan Li, Qingyang Wu, and Yong~Jae Lee.
\newblock Visual instruction tuning.
\newblock \emph{Advances in neural information processing systems}, 36, 2024{\natexlab{b}}.

\bibitem[Liu et~al.(2024{\natexlab{c}})Liu, Song, Lin, Lam, Neubig, Li, and Yue]{liu2024visualwebbench}
Junpeng Liu, Yifan Song, Bill~Yuchen Lin, Wai Lam, Graham Neubig, Yuanzhi Li, and Xiang Yue.
\newblock Visualwebbench: How far have multimodal llms evolved in web page understanding and grounding?
\newblock \emph{arXiv preprint arXiv:2404.05955}, 2024{\natexlab{c}}.

\bibitem[Lu et~al.(2024)Lu, Shao, Liu, Meng, Li, Chen, Huang, Zhang, Qiao, and Luo]{lu2024gui}
Quanfeng Lu, Wenqi Shao, Zitao Liu, Fanqing Meng, Boxuan Li, Botong Chen, Siyuan Huang, Kaipeng Zhang, Yu Qiao, and Ping Luo.
\newblock Gui odyssey: A comprehensive dataset for cross-app gui navigation on mobile devices.
\newblock \emph{arXiv preprint arXiv:2406.08451}, 2024.

\bibitem[Rawles et~al.(2024)Rawles, Li, Rodriguez, Riva, and Lillicrap]{rawles2024androidinthewild}
Christopher Rawles, Alice Li, Daniel Rodriguez, Oriana Riva, and Timothy Lillicrap.
\newblock Androidinthewild: A large-scale dataset for android device control.
\newblock \emph{Advances in Neural Information Processing Systems}, 36, 2024.

\bibitem[Sun et~al.(2022)Sun, Chen, Chen, Dai, Zhu, and Yu]{sun2022meta}
Liangtai Sun, Xingyu Chen, Lu Chen, Tianle Dai, Zichen Zhu, and Kai Yu.
\newblock Meta-gui: Towards multi-modal conversational agents on mobile gui.
\newblock \emph{arXiv preprint arXiv:2205.11029}, 2022.

\bibitem[Team et~al.(2023)Team, Anil, Borgeaud, Wu, Alayrac, Yu, Soricut, Schalkwyk, Dai, Hauth, et~al.]{team2023gemini}
Gemini Team, Rohan Anil, Sebastian Borgeaud, Yonghui Wu, Jean-Baptiste Alayrac, Jiahui Yu, Radu Soricut, Johan Schalkwyk, Andrew~M Dai, Anja Hauth, et~al.
\newblock Gemini: a family of highly capable multimodal models.
\newblock \emph{arXiv preprint arXiv:2312.11805}, 2023.

\bibitem[Touvron et~al.(2023{\natexlab{a}})Touvron, Lavril, Izacard, Martinet, Lachaux, Lacroix, Rozi{\`e}re, Goyal, Hambro, Azhar, et~al.]{touvron2023llama}
Hugo Touvron, Thibaut Lavril, Gautier Izacard, Xavier Martinet, Marie-Anne Lachaux, Timoth{\'e}e Lacroix, Baptiste Rozi{\`e}re, Naman Goyal, Eric Hambro, Faisal Azhar, et~al.
\newblock Llama: Open and efficient foundation language models.
\newblock \emph{arXiv preprint arXiv:2302.13971}, 2023{\natexlab{a}}.

\bibitem[Touvron et~al.(2023{\natexlab{b}})Touvron, Martin, Stone, Albert, Almahairi, Babaei, Bashlykov, Batra, Bhargava, Bhosale, et~al.]{touvron2023llama2}
Hugo Touvron, Louis Martin, Kevin Stone, Peter Albert, Amjad Almahairi, Yasmine Babaei, Nikolay Bashlykov, Soumya Batra, Prajjwal Bhargava, Shruti Bhosale, et~al.
\newblock Llama 2: Open foundation and fine-tuned chat models.
\newblock \emph{arXiv preprint arXiv:2307.09288}, 2023{\natexlab{b}}.

\bibitem[Wang et~al.(2024{\natexlab{a}})Wang, Xu, Ye, Yan, Shen, Zhang, Huang, and Sang]{wang2024mobile}
Junyang Wang, Haiyang Xu, Jiabo Ye, Ming Yan, Weizhou Shen, Ji Zhang, Fei Huang, and Jitao Sang.
\newblock Mobile-agent: Autonomous multi-modal mobile device agent with visual perception.
\newblock \emph{arXiv preprint arXiv:2401.16158}, 2024{\natexlab{a}}.

\bibitem[Wang et~al.(2024{\natexlab{b}})Wang, Bai, Tan, Wang, Fan, Bai, Chen, Liu, Wang, Ge, et~al.]{wang2024qwen2}
Peng Wang, Shuai Bai, Sinan Tan, Shijie Wang, Zhihao Fan, Jinze Bai, Keqin Chen, Xuejing Liu, Jialin Wang, Wenbin Ge, et~al.
\newblock Qwen2-vl: Enhancing vision-language model's perception of the world at any resolution.
\newblock \emph{arXiv preprint arXiv:2409.12191}, 2024{\natexlab{b}}.

\bibitem[Wu et~al.(2023)Wu, Wang, Shen, Peng, Nichols, and Bigham]{wu2023webui}
Jason Wu, Siyan Wang, Siman Shen, Yi-Hao Peng, Jeffrey Nichols, and Jeffrey~P Bigham.
\newblock Webui: A dataset for enhancing visual ui understanding with web semantics.
\newblock In \emph{Proceedings of the 2023 CHI Conference on Human Factors in Computing Systems}, pages 1--14, 2023.

\bibitem[Wu et~al.(2024)Wu, Han, Ding, Weng, Liu, Yao, Yu, and Kong]{wu2024copilot}
Zhiyong Wu, Chengcheng Han, Zichen Ding, Zhenmin Weng, Zhoumianze Liu, Shunyu Yao, Tao Yu, and Lingpeng Kong.
\newblock Os-copilot: Towards generalist computer agents with self-improvement.
\newblock \emph{arXiv preprint arXiv:2402.07456}, 2024.

\bibitem[Xu et~al.(2024)Xu, Yao, Guo, Cui, Ni, Ge, Chua, Liu, Sun, and Huang]{xu2024llavauhd}
Ruyi Xu, Yuan Yao, Zonghao Guo, Junbo Cui, Zanlin Ni, Chunjiang Ge, Tat-Seng Chua, Zhiyuan Liu, Maosong Sun, and Gao Huang.
\newblock Llava-uhd: an lmm perceiving any aspect ratio and high-resolution images.
\newblock \emph{arXiv preprint arXiv:2403.11703}, 2024.

\bibitem[You et~al.(2024)You, Zhang, Schoop, Weers, Swearngin, Nichols, Yang, and Gan]{you2024ferretui}
Keen You, Haotian Zhang, Eldon Schoop, Floris Weers, Amanda Swearngin, Jeffrey Nichols, Yinfei Yang, and Zhe Gan.
\newblock Ferret-ui: Grounded mobile ui understanding with multimodal llms.
\newblock \emph{arXiv preprint arXiv:2404.05719}, 2024.

\bibitem[Zhang et~al.(2024)Zhang, Li, He, Zhang, Qiao, Qin, Ma, Kang, Lin, Rajmohan, et~al.]{zhang2024ufo}
Chaoyun Zhang, Liqun Li, Shilin He, Xu Zhang, Bo Qiao, Si Qin, Minghua Ma, Yu Kang, Qingwei Lin, Saravan Rajmohan, et~al.
\newblock Ufo: A ui-focused agent for windows os interaction.
\newblock \emph{arXiv preprint arXiv:2402.07939}, 2024.

\bibitem[Zhang and Zhang(2023)]{zhang2023you}
Zhuosheng Zhang and Aston Zhang.
\newblock You only look at screens: Multimodal chain-of-action agents.
\newblock \emph{arXiv preprint arXiv:2309.11436}, 2023.

\bibitem[Zheng et~al.(2024)Zheng, Huang, Xue, Wang, An, and Yan]{zheng2024agentstudio}
Longtao Zheng, Zhiyuan Huang, Zhenghai Xue, Xinrun Wang, Bo An, and Shuicheng Yan.
\newblock Agentstudio: A toolkit for building general virtual agents.
\newblock \emph{arXiv preprint arXiv:2403.17918}, 2024.

\bibitem[Zhou et~al.(2023)Zhou, Xu, Zhu, Zhou, Lo, Sridhar, Cheng, Ou, Bisk, Fried, et~al.]{zhou2023webarena}
Shuyan Zhou, Frank~F Xu, Hao Zhu, Xuhui Zhou, Robert Lo, Abishek Sridhar, Xianyi Cheng, Tianyue Ou, Yonatan Bisk, Daniel Fried, et~al.
\newblock Webarena: A realistic web environment for building autonomous agents.
\newblock \emph{arXiv preprint arXiv:2307.13854}, 2023.

\end{thebibliography}


\begin{thebibliography}{0}
\providecommand{\natexlab}[1]{#1}
\providecommand{\url}[1]{\texttt{#1}}
\expandafter\ifx\csname urlstyle\endcsname\relax
  \providecommand{\doi}[1]{doi: #1}\else
  \providecommand{\doi}{doi: \begingroup \urlstyle{rm}\Url}\fi

\end{thebibliography}
}

\end{document}